\documentclass[conference, final]{IEEEtran}
\IEEEoverridecommandlockouts
\usepackage{amsmath,amssymb,amsfonts}
\usepackage{algorithmic}
\usepackage{graphicx}
\usepackage{textcomp}
\usepackage{xcolor}
\usepackage{bm}
\usepackage{subfigure}
\usepackage{hyperref}
\usepackage{breqn}

\usepackage{multirow}
\usepackage{amsmath}
\usepackage{amssymb}
\usepackage{caption}
\usepackage{tabularx}
\usepackage{makecell}
\usepackage{color}
\usepackage{hyperref}
\usepackage{graphicx}

\def\BibTeX{{\rm B\kern-.05em{\sc i\kern-.025em b}\kern-.08em
    T\kern-.1667em\lower.7ex\hbox{E}\kern-.125emX}}
\begin{document}

\title{Trading-off Mutual Information on Feature Aggregation for Face Recognition}

\author{
    Mohammad Akyash, Ali Zafari, Nasser M. Nasrabadi \\
    Deptartment of Computer Science \& Electrical Engineering, West Virginia University, WV USA\\

    {
    \tt\small \{\href{mailto:ma00098@mix.wvu.edu}{ma00098},\href{mailto:az00004@mix.wvu.edu}{az00004}\}@mix.wvu.edu,\{\href{mailto:nasser.nasrabadi@mail.wvu.edu}{nasser.nasrabadi}\}@mail.wvu.edu
    }\\

}

% \author{
%     Ali Zafari$^\dag$, Atefeh Khoshkhahtinat$^\dag$, Piyush M. Mehta$^\ddag$, Nasser M. Nasrabadi$^\dag$, Barbara J. Thompson$^\S$,\\
%     Michael S. F. Kirk$^\S$, Daniel da Silva$^\S$\\
%     $^\dag$Dept. of Computer Science \& Electrical Engineering, West Virginia University, WV USA\\
%     $^\ddag$Dept. of Mechanical \& Aerospace Engineering, West Virginia University, WV USA\\
%     $^\S$NASA Goddard Space Flight Center, MD USA\\
%     {
%     \tt\small \{\href{mailto:az00004@mix.wvu.edu}{az00004},\href{mailto:ak00043@mix.wvu.edu}{ak00043}\}@mix.wvu.edu,\{\href{mailto:piyush.mehta@mail.wvu.edu}{piyush.mehta},\href{mailto:nasser.nasrabadi@mail.wvu.edu}{nasser.nasrabadi}\}@mail.wvu.edu
%     }\\
%     {
%     \tt\small \{\href{mailto:barbara.j.thompson@nasa.gov}{barbara.j.thompson},\href{mailto:michael.s.kirk@nasa.gov}{michael.s.kirk},\href{mailto:daniel.e.dasilva@nasa.gov}{daniel.e.dasilva}\}@nasa.gov
    
%     }
% }

% \IEEEpubid{\begin{minipage}{\textwidth}\ \\[12pt]
%  978-1-6654-6283-9/22/\$31.00 \copyright 2022 IEEE\\ 
%  DOI 10.1109/ICMLA55696.2022.00035
% \end{minipage}} 

\maketitle

\begin{abstract}

Despite the advances in the field of Face Recognition (FR), the precision of these methods is not yet sufficient. To improve the FR performance, this paper proposes a technique to aggregate the outputs of two state-of-the-art (SOTA) deep FR models, namely ArcFace and AdaFace. In our approach, we leverage the transformer attention mechanism to exploit the relationship between different parts of two feature maps. By doing so, we aim to enhance the overall discriminative power of the FR system.
One of the challenges in feature aggregation is the effective modeling of both local and global dependencies. Conventional transformers are known for their ability to capture long-range dependencies, but they often struggle with modeling local dependencies accurately. To address this limitation, we augment the self-attention mechanism to capture both local and global dependencies effectively. This allows our model to take advantage of the overlapping receptive fields present in corresponding locations of the feature maps.
However, fusing two feature maps from different FR models might introduce redundancies to the face embedding. Since these models often share identical backbone architectures, the resulting feature maps may contain overlapping information, which can mislead the training process. To overcome this problem, we leverage the principle of Information Bottleneck to obtain a maximally informative facial representation. This ensures that the aggregated features retain the most relevant and discriminative information while minimizing redundant or misleading details.
To evaluate the effectiveness of our proposed method, we conducted experiments on popular benchmarks and compared our results with state-of-the-art algorithms. The consistent improvement we observed in these benchmarks demonstrates the efficacy of our approach in enhancing FR performance. Moreover, our model aggregation framework offers a novel perspective on model fusion and establishes a powerful paradigm for feature aggregation using transformer-based attention mechanisms.

\end{abstract}

\begin{IEEEkeywords}
Face recognition, Feature aggregation, Transformer, Cross-attention, Information bottleneck technique
\end{IEEEkeywords}

\section{\textbf{Introduction}}

\label{sec:intro}

The increased attention towards Face Recognition (FR) algorithms \cite{deng2019arcface, wang2018cosface, liu2017sphereface, kim2022adaface, talemi2023aaface} in recent years can be attributed to several factors. One of the primary catalysts has been the rising demand for reliable and efficient face recognition systems in various domains, including security \cite{rakhra2022face}, surveillance \cite{karpagam2022novel}, and identity verification \cite{BA2023}. As a result, researchers and practitioners alike have been actively exploring ways to enhance FR algorithms to meet these evolving needs.

Large-scale datasets have played a pivotal role in driving advancements in FR \cite{deng2019arcface}. These datasets comprise vast collections of annotated face images, often containing millions of samples from diverse sources. The availability of such comprehensive data allows researchers to train FR algorithms on a rich variety of facial features, appearances, and scenarios. By leveraging these datasets, FR algorithms can learn to generalize better and exhibit improved performance when faced with real-world challenges, such as variations in lighting conditions, poses, expressions, and occlusions.
In addition to large-scale datasets, novel loss functions have been instrumental in improving FR performance \cite{liu2016large}. Loss functions define the objective that FR algorithms aim to optimize during training. Traditional loss functions, such as the softmax loss, have been enhanced or replaced with more sophisticated alternatives. For instance, the triplet loss \cite{hoffer2015deep} and its variants facilitate the learning of discriminative feature representations by encouraging closer proximity for images of the same identity and pushing images of different identities further apart in the embedding space. Other loss functions, such as center loss \cite{wen2016discriminative}, focus on minimizing the intra-class variations while emphasizing inter-class separability. Most current FR methods (e.g., SphereFace \cite{liu2017sphereface}, CosFace \cite{wang2018cosface}, and ArcFace \cite{deng2019arcface}) focus on applying a margin penalty to the Softmax loss function to allow the network to extract more discriminative features. Recently, AdaFace \cite{kim2022adaface} proposed a new loss function that considers image quality during the training process and emphasizes on recognizable low quality and high quality images. 

Moreover, the development of new network architectures has significantly contributed to the progress in FR performance. Convolutional neural networks (CNNs) have revolutionized FR by effectively capturing facial features and patterns through hierarchical layers. Researchers have proposed various architectures, such as VGGNet \cite{simonyan2014very}, ResNet \cite{he2016deep}, InceptionNet \cite{szegedy2015going}, and more recently, efficient models like MobileNet \cite{howard2017mobilenets} and EfficientNet \cite{tan2019efficientnet}, each designed to extract increasingly informative representations from face images.
It is worth mentioning that advancements in hardware have also played a crucial role in facilitating the progress of FR algorithms. The availability of powerful GPUs, TPUs, and other specialized hardware accelerators has enabled researchers to train larger and more complex models efficiently. This computational power has expedited the experimentation process and allowed for more extensive exploration of network architectures, hyperparameters, and training techniques. Consequently, FR algorithms have benefited from faster training times, accelerated inference speeds, and the ability to handle large-scale datasets effectively.

Despite the progress of in the field of FR, performance of the models is still not satisfactory. \textit{Deep ensemble models} \cite{ganaie2022ensemble} mix the outputs of several independently trained methods to improve the generalization capability of the overall combination. Such ensemble models may significantly increase the accuracy of a single classifier in predicting unknown samples with high flexibility. In this paper, we exploit the transformer attention mechanism to fuse two identical networks trained with ArcFace \cite{deng2019arcface} and AdaFace \cite{kim2022adaface} loss functions.

Transformers, originally introduced for machine translation, have demonstrated exceptional performance across a wide range of natural language processing (NLP) tasks \cite{vaswani2017attention}. However, their potential extends beyond NLP, as exemplified by the Vision Transformer (ViT) \cite{dosovitskiy2020image}, which utilizes self-attention mechanisms for image recognition. ViTs have gained significant popularity and have been successfully applied to various computer vision tasks such as image classification \cite{chen2021crossvit}, compression \cite{zafari2023frequency}, object detection \cite{heo2021rethinking}, and video processing \cite{neimark2021video}.
In the field of computer vision \cite{babic2021image}, feature fusion plays a pivotal role in enhancing model accuracy by combining information from multiple sensors and modalities, resulting in more robust and comprehensive analysis \cite{alipour2023multimodal}. While transformers have been employed for feature fusion in tasks like image processing \cite{zhang2022transformer,zhou2021deep}, it is important to recognize that their main weakness lies in their inherent focus on modeling long-range dependencies between different components of visual data \cite{yuan2021incorporating}. However, when it comes to feature aggregation, preserving local dependencies becomes vital, as there is often a shared receptive field between corresponding regions in feature maps.
To address this limitation and enable effective feature aggregation, we propose a network that combines self-attention and cross-attention techniques. By leveraging these mechanisms, our network can aggregate feature maps both globally and locally. This approach allows us to effectively capture and combine both the local and global interactions between two feature maps, taking into account the specific characteristics and dependencies of the visual data.
By incorporating self-attention and cross-attention techniques, our proposed network enhances the feature aggregation process by explicitly considering local dependencies, which are crucial for accurate representation learning. This enables our method to achieve more comprehensive and informative feature representations, leading to improved performance in various computer vision tasks, including face recognition.

The Information Bottleneck (IB) principle, introduced by \cite{tishby2000information}, highlights the trade-off between learning a compact representation and achieving satisfactory prediction performance \cite{mohamadi2023more}. When combining two deep models, there is a risk of introducing redundancy into the ensemble model's output, potentially misleading the training process. To address this concern, we leverage the IB principle to obtain a compressed yet informative aggregated representation of the two model features.
To achieve this, we incorporate a regularization term into the loss function that suppresses irrelevant information in the aggregated representation. This regularization encourages the network to focus on capturing essential and discriminative features while disregarding redundant or irrelevant details. By explicitly incorporating the IB principle into our model, we aim to strike a balance between compactness and performance, resulting in a more effective and efficient representation.
The original IB method involves computationally expensive calculations of mutual information between the input, latent representation, and output. To mitigate this challenge, we adopt the concept of Variational Information Bottleneck (VIB) \cite{alemi2016deep}. The VIB approach approximates the mutual information by introducing a variational lower bound, enabling more efficient computation and scalability to large-scale datasets.
We evaluate our proposed model on a range of benchmark datasets, including AgeDB \cite{moschoglou2017agedb}, CFP-FP \cite{sengupta2016frontal}, CPLFW \cite{zheng2018cross}, CALFW \cite{zheng2017cross}, LFW \cite{huang2008labeled}, IJB-B \cite{whitelam2017iarpa}, and IJB-C \cite{maze2018iarpa}. Through these evaluations, we demonstrate significant improvements in performance compared to state-of-the-art (SOTA) algorithms, validating the effectiveness of our approach.
By leveraging the IB principle and adopting the VIB framework, we address the challenge of redundancy in ensemble models while achieving a compressed and informative aggregated representation. The experimental results across various datasets underscore the superiority of our method, highlighting its potential for advancing the field of face recognition and outperforming existing state-of-the-art techniques.

To sum up, the contributions of this work are as follows:

\begin{enumerate}
  \item    A novel local-global transformer-based neural network is proposed to aggregate the output features of the ArcFace and AdaFace methods.
  
  \item By employing the information bottleneck principle, we declare that the final output feature embedding is refined and does not have redundancies. In other words, our loss function guides the fusion network to suppress irrelevant information in the representation.
  
  \item To demonstrate the efficacy of the proposed method, we perform extensive experiments on publicly-available datasets. Results confirm our technique performs well across various benchmarks.
\end{enumerate}

The remainder of this paper is as follows: In Section \ref{sec:method}, we elaborate on our method. In Section \ref{sec:experiments}, we evaluate our model and compare it to the SOTA methods. Finally, in Section \ref{sec:conclution} we conclude the paper.

\section{\textbf{Related Works}}

\subsection{\textbf{Face Recognition methods}}

In face recognition, a common margin-based loss functions aim to improve the discriminative power of the learned features by explicitly enforcing a margin between different classes in the feature space. It is commonly used in face recognition tasks to enhance the inter-class separability. In this subsection we introduce the recent advances in margin-based loss functions.

SphereFace \cite{liu2017sphereface} introduces a novel angle-based softmax loss that incorporates a margin function to enhance the discriminative power of the learned features. The margin function is designed to ensure that the features belonging to different classes are well separated in the angular space. The angle-based softmax loss used in SphereFace can be defined as:

\begin{equation}
    \resizebox{0.45\textwidth}{!}{$\displaystyle
    L_{\text{sphere}} = -\log\left(\frac{{\exp(s(\cos(\theta_{yi} - m)))}}{{\exp(s(\cos(\theta_{yi} - m))) + \sum_{j \neq yi} \exp(s\cos(\theta_j))}}\right),
    $}
\end{equation}

where $L_{sphere}$ is the loss function, s is a scaling factor, $\theta_{yi_{m}}$ is the angle between the input feature and the weight vector of the ground truth class $yi$ after applying a margin $m$, and $\theta_{j}$ is the angle between the input feature and the weight vector of class $j$. The margin $m$ is introduced to increase the angular separation between different classes. CosFace \cite{wang2018cosface} focuses on enhancing the margin-based loss by incorporating the cosine similarity metric. The margin function used in CosFace is designed to increase the angular separation between different classes in the feature space. The CosFace loss function can be defined as:

\begin{equation}
\resizebox{0.45\textwidth}{!}{$\displaystyle
    L_{\text{cos}} = -\log\left(\frac{{\exp(s(\cos(\theta_{yi}) - m)))}}{{\exp(s(\cos(\theta_{yi}) - m))) + \sum_{j \neq yi} \exp(s\cos(\theta_j))}}\right),
    $}
\end{equation}

where $L_{cos}$ is the loss function, $s$ is a scaling factor, $\theta_{yi}$ is the angle between the input feature and the weight vector of the ground truth class $yi$, and $\theta_j$ is the angle between the input feature and the weight vector of class $j$. The margin $m$ is added to increase the angular margin between classes. ArcFace \cite{deng2019arcface} builds upon the concept of using a margin function within the softmax loss to improve the discriminative capacity of the learned features. The margin function in ArcFace is designed to enforce large angular separations between classes in the feature space. The ArcFace loss function can be defined as:

\begin{equation}
\resizebox{0.45\textwidth}{!}{$\displaystyle
    L_{\text{arc}} = -\log\left(\frac{{\exp(s(\cos(\theta_{yi} + m)))}}{{\exp(s(\cos(\theta_{yi} + m))) + \sum_{j \neq yi} \exp(s\cos(\theta_j))}}\right),
    $}
\end{equation}

where $L_{arc}$ is the loss function, $s$ is a scaling factor, $\theta_{yi}$ is the angle between the input feature and the weight vector of the ground truth class $yi$, and $\theta_j$ is the angle between the input feature and the weight vector of class $j$. The margin $m$ is added to increase the angular separation between classes.

AdaFace \cite{kim2022adaface} introduces an adaptive margin-based approach that dynamically adjusts the margin for each training sample, leading to improved discriminability. The margin adaptation process in AdaFace aims to handle intra-class variations by assigning larger margins to challenging samples and smaller margins to easier ones. The AdaFace loss function can be defined as:

\begin{equation}
\resizebox{0.45\textwidth}{!}{$\displaystyle
    L_{\text{ada}} = -\log\left(\frac{{\exp(s(\cos(\theta_{yi} + m_i)))}}{{\exp(s(\cos(\theta_{yi} + m_i))) + \sum_{j \neq yi} \exp(s\cos(\theta_j))}}\right),
    $}
\end{equation}

Where $L_{ada}$ is the loss function, $s$ is a scaling factor, $\theta_{yi}$ is the angle between the input feature and the weight vector of the ground truth class $yi$, $\theta_j$ is the angle between the input feature and the weight vector of class $j$, and $m_{i}$ is the dynamically adjusted margin for each training sample. The margin $m_{i}$ is computed based on the difficulty or intra-class variations of the sample.
By employing margin-based loss functions, including the specific forms used in SphereFace, CosFace, ArcFace, and the adaptive margin approach in AdaFace, these techniques aim to enhance the discriminative power of face recognition models and improve their performance in challenging scenarios.

 One of the key advantages of these methods is their ability to enhance the discriminative power of the learned features. By incorporating margin functions within the softmax loss, these techniques effectively increase the angular separations between different classes in the feature space. This leads to better discrimination between individuals, resulting in more accurate face recognition. Another advantage of these methods is their robustness to variations commonly encountered in face recognition, such as pose variations, lighting conditions, and occlusions. By incorporating margin functions and angular constraints, these techniques encourage the learned features to be less affected by variations, resulting in improved robustness and generalization capabilities.

\subsection{\textbf{Ensemble Learning methods}}

\begin{figure*}[t]
\scalebox{1.55}{\includegraphics{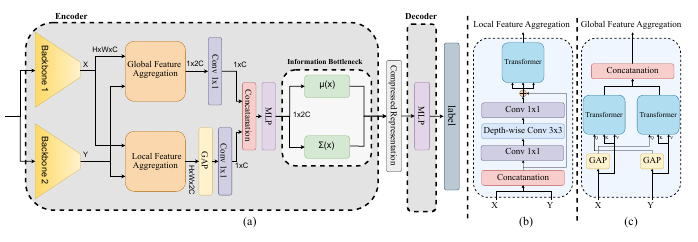}}
\centering
\caption{(a) The overall architecture of the proposed method. The input image is fed to two pre-trained FR backbones to obtain two feature maps $X, Y\in{R^{H\times W\times C}}$. Next, the feature maps are aggregated with the introduced global and local transformer-based modules and then, we exploit the IB technique to obtain a compressed representation of the image. (b) The local feature aggregation module. (c) The global aggregation module.}
\label{overall_framework}
\end{figure*}

In machine learning, an ensemble model is a technique that combines multiple individual models to make predictions. The idea behind ensemble models is that the combination of several weak models can result in a stronger and more accurate predictor. In this subsection, we review some of the well-known ensemble learning methods. Random Forest is a popular ensemble method introduced by \cite{breiman2001random}. It combines multiple decision trees, where each tree is trained on a random subset of the data and features. Random Forest has been widely used in various domains due to its robustness and ability to handle high-dimensional data.
 Gradient Boosting Machines (GBM) \cite{friedman2001greedy} is another powerful ensemble technique that builds an ensemble of weak prediction models, typically decision trees, in a sequential manner. Each model is trained to correct the mistakes made by the previous models. Notable GBM implementations include XGBoost \cite{chen2015xgboost} and LightGBM \cite{ke2017lightgbm}, which have gained significant popularity due to their efficiency and performance. Deep ensembles, as discussed previously, combine multiple deep learning models to form an ensemble. These models are typically deep neural networks trained independently and combined using techniques such as averaging or voting. Deep ensembles have been applied to various domains, including image classification, natural language processing, and reinforcement learning.
Bayesian Model Averaging (BMA) \cite{hoeting1999bayesian} is a probabilistic ensemble approach that assigns weights to individual models based on their performance on a validation set. Bayesian techniques are used to estimate the weights, and the final prediction is obtained by averaging the predictions of all models weighted by their probabilities.
Stochastic Weight Averaging (SWA) is a recent ensemble technique proposed by \cite{izmailov2018averaging}. It involves averaging the weights of multiple models during training rather than their predictions. This method has been shown to improve generalization and robustness of deep learning models.

 \section{\textbf{Method}}
\label{sec:method}

% \subsection{Multi-Head self Attention}

% The base block of our proposed network in Multi-Head self Attention. Single-head self-attention is expanded by multi-head attention to include numerous intricate relationships between different input tokens. An integral element of Transformers, the self-attention mechanism assigns a paired attention score between every two patch tokens in terms of global contextual information, methodically simulating the connection between every token of the input.

\begin{figure}[t]
\scalebox{1.3}{\includegraphics[angle=270]{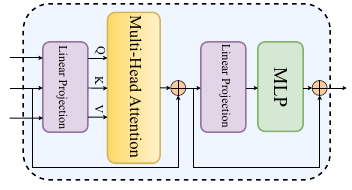}}
\centering
\caption{The transformer encoder architecture, the base block of our proposed local and global feature aggregation modules.}
\label{transformer}
\end{figure}

Figure \ref{overall_framework}(a) illustrates the comprehensive architecture of our proposed method. To begin, we extract feature maps from the final convolutional layer of each pre-trained face recognition (FR) backbone. These feature maps serve as input to two parallel modules within our approach, enabling simultaneous local and global information aggregation.
To facilitate this aggregation process, we employ a transformer encoder architecture as depicted in Figure \ref{transformer}. The transformer encoder acts as the fundamental building block of our feature aggregation modules, allowing for efficient capturing and integration of local and global facial information.
Following the feature aggregation step, we concatenate the outputs from the local and global modules and leverage the Information Bottleneck (IB) technique. This technique enables us to achieve a compressed representation that retains the essential discriminative information while removing redundancies, optimizing the overall performance of the face recognition model.
Finally, we decode the compressed representation to obtain the corresponding labels, allowing for accurate classification. During the inference phase, the compressed representation is utilized for 1:1 face verification, enabling efficient and reliable matching between pairs of face images.
By adopting this comprehensive architecture, our method effectively combines local and global information, leverages the power of transformers for feature aggregation, incorporates the benefits of the Information Bottleneck principle for compression, and ultimately enables accurate face recognition and verification tasks.

\subsection{\textbf{Information Bottleneck method}}

In our loss function, we have used the IB principle \cite{tishby2000information} to achieve a compressed and informative fused representation of the images. In a classification task, we need to learn a representation that is maximally compressed with regard to the input and maximally informative about the output. The IB principle is defined below:
\begin{equation}
\mathcal{L}_{IB}(\theta) = \beta I(\hat{X},X;\theta) - I(\hat{X},Y;\theta),
\end{equation}
\noindent where $I(.,.)$ denotes the mutual information, and $X$, $\hat{X}$, and $Y$ represent the input, bottleneck representation, and corresponding labels, respectively.

\subsubsection{\textbf{Variational Information Bottleneck}}

The main drawback of the IB principle is that the computation of mutual information is cumbersome, especially for continuous and high-dimensional variables. Recently, remarkable improvements have allowed the computation of MI in an efficient manner \cite{alemi2016deep,belghazi2018mutual}. In \cite{alemi2016deep} a variational bound is presented to approximate the IB objective. This bound is defined below:

\begin{dmath}
\mathcal{L}_{IB}(\theta) = \beta I(\hat{X},X;\theta) - 
I(\hat{X},Y;\theta) \le \beta\int_{}^{}p(x)p_{\theta}
(\hat{x}|x)log\frac{p_{\theta}(\hat{x}|x)}{r(\hat{x})}dxd\hat{x} - \int_{}^{}p(x)p(y|x)p_{\theta}(x|\hat{x})log_{q_\phi}(y|\hat{x})dxdyd\hat{x}
\end{dmath},
\noindent where $p_{\theta}(\hat{x}|x)$ is the estimation for the posterior probability, $r(\hat{x})$ is a normal distribution and $q_{\theta}(y|\hat{x})$ is the estimation of distribution $Y$. The loss function is then defined below:
\begin{dmath}
\label{vib}
\mathcal{L}_{VIB}(\theta, \phi) = \beta \mathbb{E}_{x}{[KL(p_{\theta}(\hat{x}|x), r(\hat{x}))]} + \mathbb{E}_{\hat{x} \sim p_{\theta}(\hat{x}|x)}{[-log({q_\phi}(y|\hat{x}))]}
\end{dmath},
\noindent where $D_{KL}(.,.)$ denotes the \textit{Kullback-Leibler divergence}.

\subsection{\textbf{Attention-Based Fusion Architecture}}

\subsubsection{\textbf{Local Feature Aggregation}}

While a transformer attention mechanism excels at capturing long-range dependencies across input tokens, it does not inherently emphasize the interaction between tokens within a local region. In our feature fusion approach, where we compute feature maps using two identical backbones, the pixels within a local region of the feature maps share common receptive fields in the original image. Therefore, it is crucial to design a model that can effectively capture the relationships between elements in corresponding local positions of the two feature maps.

To address this requirement, we propose a variation of the transformer architecture that specifically focuses on modeling the local context between different parts of the input features. Figure \ref{overall_framework}(b) illustrates the architecture of our local feature aggregation module. Inspired by the work of \cite{yuan2021incorporating}, we modify the conventional transformer to enhance its capability to capture local context.
To enable the transformer to effectively capture local context, we begin by concatenating the two feature maps, denoted as $X$ and $Y$, resulting in a composite feature map $F$ of shape $H \times W \times 2C$. We then apply a sequence of operations, including a $Conv(1 \times 1)$, a depth-wise convolution, and another $Conv(1 \times 1)$, followed by adding $F$ to the output. This series of operations is designed to establish local context fusion within the transformer, a capability that the conventional transformer lacks.
Subsequently, the resulting feature map is flattened to $F^\prime$ of shape $HW \times C$ and fed into the transformer module. In this configuration, the number of tokens is $HW$, each having a size of $C$. By incorporating the depth-wise convolution, we aim to effectively capture and integrate local context within the transformer, enabling it to model the relationships between elements in local regions.
It is important to note that the introduction of the depth-wise convolution does not significantly increase the computational complexity compared to the original transformer. This is due to the low computational overhead associated with depth-wise convolutions. Therefore, our modified architecture remains computationally efficient while effectively capturing local context and enhancing feature fusion.
By leveraging the local feature aggregation module, we enable our model to capture both global and local interactions within the feature maps. This comprehensive understanding of relationships between elements contributes to more accurate and robust feature fusion, ultimately enhancing the performance of our approach in face recognition.

\subsubsection{\textbf{Global Feature Aggregation}}

While Transformers are known for their ability to model global interactions among tokens, they typically consider the relationship of a single token with other tokens at a time, neglecting the potential for fully utilizing the interaction among all elements within a feature. To enhance performance and enable a more comprehensive understanding of the global relations between elements in two features, we modify the Transformer network accordingly.

The global feature fusion module, depicted in Figure \ref{overall_framework}(c), plays a crucial role in this enhancement. The input to this module consists of two feature maps, denoted as $X$ and $Y$, both of shape $H \times W \times C$. To facilitate interaction between the feature maps, we apply Global Average Pooling (GAP) to each feature map, resulting in two feature vectors, $G_X$ and $G_Y$, both of shape $1 \times C$.
To enable the feature maps to interact with each other, we utilize the feature vector $G_Y$ as the query of one Transformer, and $G_X$ as the query of another Transformer. The keys and values for the Transformers are derived from $Y$ and $X$, respectively. The result of applying a $Conv(1 \times 1)$ operation to $GAP(Y)$, $X$, and $Y$ produces $K$ and $V$, both of shape $HW \times C$.
The multi-head attention mechanism is then applied, resulting in $Z_1$, an output of shape $1 \times C$. Following the attention step, the output passes through the "Add and Norm," "FeedForward," and "Add and Norm" layers. These operations further refine the features and contribute to the final representations, resulting in two feature vectors, $F_1$ and $F_2$, both of shape $1 \times C$.
The two attended feature vectors, $F_1$ and $F_2$, are concatenated to form a resulting feature vector, denoted as $F$, of shape $2 \times C$. This fusion allows for the integration of the attended information from both feature maps, capturing a richer representation of the global relations between the elements.
Importantly, in this fusion process, the query of each transformer module is the average of all the pixels in a channel, enabling a global perspective for attention. Consequently, the values are weighted based on the interaction between the "globally designed" query and the key. This attention mechanism ensures that the interaction among the elements is effectively captured and utilized to enhance the overall representation.

\subsection{\textbf{Objective Function}}

The VIB objective function serves as the loss function for training our network. In the literature of variational autoencoders, the encoder and decoder are denoted as $p_{\theta}(\hat{x}|x)$ and $q_{\phi}(y|\hat{x})$ respectively. To align with this convention, we rewrite Equation \ref{vib} as follows:

\begin{equation}
\mathcal{L}_{VIB}(\theta, \phi) = \beta \mathcal{L}_{encoder}(\theta) + \mathcal{L}_{decoder}(\phi).
\label{loss}
\end{equation}

In our network architecture, the backbones and the global-local feature aggregation components collectively function as our encoder. On the other hand, the decoder consists of a fully connected layer that connects the bottleneck representation $\hat{x}$ with the corresponding labels.
Within Equation \ref{loss}, the $\mathcal{L}_{decoder}$ term represents the cross-entropy loss, which measures the dissimilarity between the predicted labels and the true labels. On the other hand, the $\mathcal{L}_{encoder}$ term acts as a regularization term, imposing constraints on the network to encourage the removal of redundant information from the learned representation $\hat{x}$. This regularization facilitates a more compact and informative representation.
During the training process, the backbones remain fixed, so the trainable weights $\theta$ correspond to the fusion network. Additionally, the weights $\phi$ represent those of the fully connected layer. The hyperparameter $\beta$ determines the extent of compression applied to the learned representation.
To approximate the parameters of $p_{\theta}(z|x)$, we make use of the approximations $\mu(x)$ and $\Sigma(x)$. During training, the compressed representation $\hat{x}$ is sampled from $p_{\theta}(z|x)$, while during inference, we utilize $\mu(x)$ as the representation for the input image.
By formulating our loss function in this manner and incorporating the VIB objective, we enable the network to simultaneously optimize the decoder for accurate classification and the encoder for efficient representation compression. This framework allows us to achieve a balance between preserving predictive information and eliminating redundancies, ultimately enhancing the face recognition performance of our model.

\section{\textbf{Experimental Setting and Results}}
\label{sec:experiments}
% Please add the following required packages to your document preamble:
% \usepackage{multirow}
% Please add the following required packages to your document preamble:
% \usepackage{multirow}
% Please add the following required packages to your document preamble:
% \usepackage{ultirow}

\begin{table*}[]
\caption{The results of our proposed method are compared to the SOTA methods for the 1:1 face verification task.}

\scalebox{1.3}{
\begin{tabular}{c|ccccc|cc}

\Xhline{2\arrayrulewidth}\hline
\multirow{2}{*}{Method} & \multicolumn{5}{c|}{High Quality}                                                  & \multicolumn{2}{c}{Mixed Quality} \\ \cline{2-8} 
                        & LFW            & CFP-FP         & CPLFW          & AgeDB          & CALFW          & IJB-B           & IJB-C           \\ \hline
ArcFace \cite{deng2019arcface}                & 99.83          & 98.27          & 92.08          & 98.28          & 95.45          & 94.25           & 96.03           \\
AdaFace  \cite{kim2022adaface}                 & 99.82          & 98.49          & 93.53          & 98.05          & 96.08          & 95.67           & 96.89           \\
CurricularFace \cite{huang2020curricularface}             & 99.80          & 98.37          & 93.13          & 98.32 & 96.20          & 94.80           & 96.10           \\
MagFace  \cite{meng2021magface}             & 99.83          & 98.46          & 92.87          & 98.17          & 96.15          & 94.51           & 95.97           \\
BroadFace \cite{kim2020broadface}              & \textbf{99.85} & 98.63          & 93.17          & 98.38          & 96.20          & 94.97           & 96.37           \\
SCF-ArcFace  \cite{li2021spherical}          & 99.82          & 98.40          & 93.16          & 98.30          & 96.12          & 94.74           & 96.09           \\ \hline
Fusion + Global         & 99.83          & 98.54          & 93.46          & 98.37          & 96.15          & 95.54           & 96.75           \\
Fusion + Local          & 99.83          & 98.46          & 93.50          & 98.25          & 96.20          & 95.63           & 96.84           \\
Fusion + Global + Local         & 99.83          & 98.50          & 93.53          & 98.30          & 96.18          & 95.65           & 96.89           \\
Fusion + Global + Local + VIB & \textbf{99.85} & \textbf{98.87} & \textbf{93.78} & \textbf{98.60} & \textbf{96.25} & \textbf{95.83}  & \textbf{97.11}  \\ \hline\hline

\end{tabular}}
\label{result}
\centering
\end{table*}

\subsection{\textbf{Datasets}}

We have incorporated a segment of WebFace12M \cite{zhu2021webface260m}, comprising over 5 million facial images, into our training dataset. During the testing phase, we evaluated our model using diverse datasets with varying image qualities to assess its robustness and generalization capabilities.
For the high-quality image datasets, we utilized AgeDB \cite{moschoglou2017agedb}, CFP-FP \cite{sengupta2016frontal}, CPLFW \cite{zheng2018cross}, CALFW \cite{zheng2017cross}, LFW \cite{huang2008labeled}, IJB-B \cite{whitelam2017iarpa}, and IJB-C \cite{maze2018iarpa} datasets, which are widely recognized benchmarks within the face recognition (FR) community. These datasets are known for their well-captured and well-aligned images, providing an ideal evaluation environment for assessing the performance of FR methods.
To ensure consistent and standardized evaluation, we pre-processed each dataset using the techniques outlined in \cite{deng2020retinaface}, which includes face detection and alignment procedures. Furthermore, we followed the settings defined in \cite{deng2019arcface} to perform rescaling and alignment, ensuring fair and comparable evaluation conditions across different methods.
To align with the evaluation practices of state-of-the-art methods, we reported the 1:1 verification accuracy for the aforementioned datasets. Additionally, we presented the True Accept Rate (TAR) at a False Accept Rate (FAR) of $1e-4$, which provides a comprehensive measure of the model's performance in distinguishing genuine matches from impostor matches.
By adhering to established evaluation standards and employing a diverse range of datasets, including those with both high-quality and low-quality images such as IJB-B and IJB-C, we aim to demonstrate the effectiveness and versatility of our proposed method across various real-world scenarios.

\subsection{\textbf{Implementation details}}
For the backbones, we exploit the ResNet100, pre-trained with ArcFace \cite{deng2019arcface} and AdaFace \cite{kim2022adaface} losses with the same training dataset as ours. Using stochastic gradient descent (SGD), the entire network is trained for 24 epochs with the $\mathcal{L}_{VIB}$ loss function. The learning rate begins at 0.1 and is decreased by a factor of 10 at 10$^{th}$, 16$^{th}$, and 22$^{th}$ epochs. For the training phase, each image and its corresponding label (as a one-hot vector) are fed to the network. During the inference phase, the pair is given to the network, and the cosine distance is then computed between the representations as a metric.
For experiment, first, we evaluate only the global branch then we use the local branch, and at the end, we take advantage of both the local and global branches. In the last experiment, $\mathcal{L}_{VIB}$ is used, and the compressed representation length is $K = 512$.

\subsection{\textbf{Comparison with the SOTA methods}}

In Table \ref{result}, we exhibit the performance of our algorithm compared to the state-of-the-art techniques, and as we can see in the table, our method outperforms these algorithms. We evaluate our model with global, local, and both local and global modules, and we achieved the best results when we used both modules simultaneously. To demonstrate the effectiveness of the loss function $\mathcal{L}_{VIB}$, we conduct experiments with and without VIB, and the results indicate that by using this loss function, we can achieve better performance.
Furthermore, in the next part, we explain how tuning the hyper-parameter $\beta$ allows us to achieve the optimum performance on both high and mixed-quality datasets. The hyper-parameter $\beta$ controls the trade-off between the reconstruction fidelity and the disentanglement of the latent representations. Through careful tuning of $\beta$, we are able to strike a balance that ensures high performance across various datasets.

\subsection{\textbf{Effect of hyper-parameter $\beta$ on accuracy}} 
\begin{figure}
    
  \centering
  \scalebox{.55}{\includegraphics{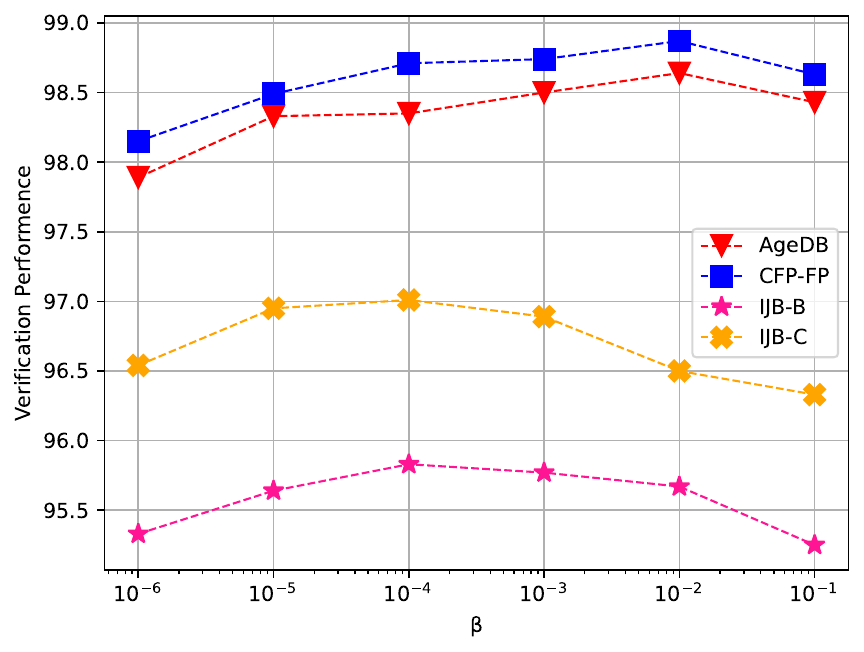}}
  \caption{The verification performance versus the hyper-parameter $\beta$. To achieve optimal performance for high quality images we need higher $\beta$ (more compression) due to the higher amount of contextual information.}
  \label{fig:qplot}
\end{figure}

We examine how the hyper-parameter $\beta$ in $\mathcal{L}_{VIB}$ affects the face recognition performance by controlling the trade-off between preserving predictive information and compression in the latent representation. A low $\beta$ value indicates a greater emphasis on preserving predictive information, while a higher $\beta$ value prioritizes compression and eliminates redundancies.
To investigate the impact of $\beta$ on performance, we trained our model with various values of this parameter. Fig. \ref{fig:qplot} illustrates the verification accuracy on the validation datasets, providing insights into the relationship between $\beta$ and performance.
Our experiments reveal that different datasets exhibit varying sensitivity to $\beta$ due to differences in image quality. For high-quality datasets such as AgeDB and CFP-FP, a higher $\beta$ value is required. These datasets contain rich contextual information, necessitating a greater degree of compression in their representations. By increasing $\beta$, we can effectively eliminate redundancies and achieve improved performance on these high-quality images.
Conversely, for lower-quality images, a lower $\beta$ value proves more effective. These images contain less contextual information and may benefit from a more informative representation that preserves a higher degree of target-related details. By reducing $\beta$, we strike a balance that allows for more preservation of relevant information in the face recognition process for lower-quality datasets.
In summary, our investigation demonstrates the importance of tuning $\beta$ in $\mathcal{L}_{VIB}$ to achieve optimal face recognition performance. The appropriate choice of $\beta$ depends on the dataset characteristics, with higher values suitable for high-quality images and lower values preferred for lower-quality images. This flexibility enables our model to adapt to different datasets and maximize performance across a range of image qualities. 

\section{\textbf{conclusion}}

In this paper, we propose a transformer-based architecture to enhance face recognition performance by aggregating the output features of two pre-trained networks. Transformers have limitations in capturing local interactions, so we divided the fusion module into local and global feature aggregation components.
To address potential redundancies in the aggregated features, we leverage the Information Bottleneck (IB) principle to achieve a maximally informative and compressed representation. We evaluate our model on various benchmarks and demonstrate its superiority over SOTA methods.
Overall, our approach effectively addresses the limitations of transformers in capturing local interactions by dividing the fusion module and leveraging the IB principle. The experimental results showcase the improved performance of our model compared to existing methods. This paper contributes to the field of face recognition by proposing an enhanced architecture for real-world scenarios.
\label{sec:conclution}

\section{\textbf{Acknowledgment}}

This research is based upon work supported by the Office of the Director of National Intelligence (ODNI), Intelligence Advanced Research Projects Activity (IARPA), via IARPA R\&D Contract No. 2022-21102100001. The views and conclusions
contained herein are those of the authors and should not be interpreted as necessarily representing the official policies or endorsements, either expressed or implied, of the ODNI, IARPA, or the U.S. Government. The U.S. Government is authorized to reproduce and distribute reprints for Governmental purposes notwithstanding any copyright annotation thereon.
\label{sec:ack}

\bibliographystyle{IEEEtran}
\bibliography{refs}
\end{document}